\documentstyle[11pt,colap]{article}
\begin{document}

\title{\vspace{-0.5in}Coordination in Tree Adjoining Grammars:\\
  Formalization and Implementation\thanks{This work is partially
    supported by NSF grant NSF-STC SBR 8920230. ARPA grant N00014-94
    and ARO grant DAAH04-94-G0426. We want to thank Nobo Komagata,
    Seth Kulick, Jong Park, James Rogers, B. Srinivas, Mark Steedman,
    and two anonymous reviewers for their valuable comments.}}

\author{Anoop Sarkar and Aravind Joshi \\
  Department~of Computer and Information Science \\
  University of Pennsylvania \\
  Philadelphia, PA 19104 \\
  {\tt \{anoop,joshi\}@linc.cis.upenn.edu}}

\maketitle
\vspace{-0.5in}
\hyphenation{L-TAGs}

\begin{abstract}
  In this paper we show that an account for coordination can be
  constructed using the derivation structures in a lexicalized Tree
  Adjoining Grammar (LTAG).  We present a notion of derivation in
  LTAGs that preserves the notion of fixed constituency in the LTAG
  lexicon while providing the flexibility needed for coordination
  phenomena. We also discuss the construction of a practical parser
  for LTAGs that can handle coordination including cases of
  non-constituent coordination.
\end{abstract}

\bibliographystyle{acl}


\newcommand{\oneoverb}[2]{\begin{array}[b]{@{}c@{}} #1 \\
\hline #2  \end{array}}

\newcommand{\twooverb}[2]{\begin{array}[b]{@{}c@{}} #1 \\
\hline \hline #2 \end{array}}

\newcommand{\oneovermodb}[3]{\begin{array}[b]{@{}c@{}}
                                   #2 \\[-1.8ex]
                                   \hbox to #1{\hrulefill} \\[-.8ex]
                                   #3 \end{array}}

\newcommand{\oneover}[2]{\begin{array}{@{}c@{}} #1 \\
\hline #2  \end{array}}

\newcommand{\twoover}[2]{\begin{array}{@{}c@{}} #1 \\
\hline \hline #2 \end{array}}

\newcommand{\oneovermod}[3]{\begin{array}{@{}c@{}}
                                   #2 \\[-1.8ex]
                                   \hbox to #1{\hrulefill} \\[-.8ex]
                                   #3 \end{array}}


\newcounter{sentencectr}
\newcounter{sentencesubctr}

\renewcommand{\thesentencectr}{(\smainform{sentencectr})}
\renewcommand{\thesentencesubctr}{\thesentencectr\ssubform{sentencesubctr}}

\newcommand{\smainform}{\arabic}
\newcommand{\ssubform}{\alph}
\newcommand{\ssubpunc}{.{}}

\newcommand{\beginsentences}{ %
\pagebreak[3] %
\begin{list}{(\thesentencectr)}
   {\usecounter{sentencesubctr}
    \setlength{\topsep}{1ex}			
    \setlength{\itemsep}{0 in}
    \setlength{\labelwidth}{0.5 in}
    \addtolength{\leftmargin}{4ex}
    \setlength{\labelsep}{.05in}
    \setlength{\parsep}{0 in}}}
\def\endsentences{\end{list}}

\newcommand{\sitem}{\renewcommand{\thesentencesubctr}{(\smainform{sentencectr}}
                    \refstepcounter{sentencectr}
     \item[(\smainform{sentencectr})\hfill]}
\newcommand{\smainitem}{\renewcommand{\thesentencesubctr
                                    }{\thesentencectr\ssubform{sentencesubctr}}
                        \setcounter{sentencesubctr}{0}
                        \refstepcounter{sentencectr}
                        \refstepcounter{sentencesubctr}
     \item[\thesentencectr\hfill\ssubform{sentencesubctr}\ssubpunc]}
\newcommand{\ssubitem}{\refstepcounter{sentencesubctr}
     \item[\hfill\ssubform{sentencesubctr}\ssubpunc]}

\makeatletter            
\newcommand{\smainlabel}[1]{{
\renewcommand{\@currentlabel}{\thesentencectr}\label{#1}}}

\newcommand{\ssublabel}[1]{{
\renewcommand{\@currentlabel}{\ssubform{sentencesubctr}}\label{#1}}}
\makeatother

\makeatother

\catcode`\@=11\relax
\newwrite\@unused
\def\typeout#1{{\let\protect\string\immediate\write\@unused{#1}}}
\typeout{psfig/tex 1.2-dvips}


\def\figurepath{./}
\def\psfigurepath#1{\edef\figurepath{#1}}

%
%
\def\@nnil{\@nil}
\def\@empty{}
\def\@psdonoop#1\@@#2#3{}
\def\@psdo#1:=#2\do#3{\edef\@psdotmp{#2}\ifx\@psdotmp\@empty \else
    \expandafter\@psdoloop#2,\@nil,\@nil\@@#1{#3}\fi}
\def\@psdoloop#1,#2,#3\@@#4#5{\def#4{#1}\ifx #4\@nnil \else
       #5\def#4{#2}\ifx #4\@nnil \else#5\@ipsdoloop #3\@@#4{#5}\fi\fi}
\def\@ipsdoloop#1,#2\@@#3#4{\def#3{#1}\ifx #3\@nnil 
       \let\@nextwhile=\@psdonoop \else
      #4\relax\let\@nextwhile=\@ipsdoloop\fi\@nextwhile#2\@@#3{#4}}
\def\@tpsdo#1:=#2\do#3{\xdef\@psdotmp{#2}\ifx\@psdotmp\@empty \else
    \@tpsdoloop#2\@nil\@nil\@@#1{#3}\fi}
\def\@tpsdoloop#1#2\@@#3#4{\def#3{#1}\ifx #3\@nnil 
       \let\@nextwhile=\@psdonoop \else
      #4\relax\let\@nextwhile=\@tpsdoloop\fi\@nextwhile#2\@@#3{#4}}
\def\psdraft{
	\def\@psdraft{0}
}
\def\psfull{
	\def\@psdraft{100}
}
\psfull
\newif\if@prologfile
\newif\if@postlogfile
\newif\if@noisy
\def\pssilent{
	\@noisyfalse
}
\def\psnoisy{
	\@noisytrue
}
\psnoisy
\newif\if@bbllx
\newif\if@bblly
\newif\if@bburx
\newif\if@bbury
\newif\if@height
\newif\if@width
\newif\if@scale
\newif\if@rheight
\newif\if@rwidth
\newif\if@clip
\newif\if@verbose
\def\@p@@sclip#1{\@cliptrue}


\def\@p@@sfile#1{\def\@p@sfile{null}%
	        \openin1=#1
		\ifeof1\closein1%
		       \openin1=\figurepath#1
			\ifeof1\typeout{Error, File #1 not found}
			\else\closein1
			    \edef\@p@sfile{\figurepath#1}%
                        \fi%
		 \else\closein1%
		       \def\@p@sfile{#1}%
		 \fi}
\def\@p@@sfigure#1{\def\@p@sfile{null}%
	        \openin1=#1
		\ifeof1\closein1%
		       \openin1=\figurepath#1
			\ifeof1\typeout{Error, File #1 not found}
			\else\closein1
			    \def\@p@sfile{\figurepath#1}%
                        \fi%
		 \else\closein1%
		       \def\@p@sfile{#1}%
		 \fi}

\def\@p@@sbbllx#1{
		\@bbllxtrue
		\dimen100=#1
		\edef\@p@sbbllx{\number\dimen100}
}
\def\@p@@sbblly#1{
		\@bbllytrue
		\dimen100=#1
		\edef\@p@sbblly{\number\dimen100}
}
\def\@p@@sbburx#1{
		\@bburxtrue
		\dimen100=#1
		\edef\@p@sbburx{\number\dimen100}
}
\def\@p@@sbbury#1{
		\@bburytrue
		\dimen100=#1
		\edef\@p@sbbury{\number\dimen100}
}
\def\@p@@sscale#1{
		\@scaletrue
		\count255=#1
   		\edef\@p@sscale{\number\count255}
}
\def\@p@@sheight#1{
		\@heighttrue
		\dimen100=#1
   		\edef\@p@sheight{\number\dimen100}
}
\def\@p@@swidth#1{
		\@widthtrue
		\dimen100=#1
		\edef\@p@swidth{\number\dimen100}
}
\def\@p@@srheight#1{
		\@rheighttrue
		\dimen100=#1
		\edef\@p@srheight{\number\dimen100}
}
\def\@p@@srwidth#1{
		\@rwidthtrue
		\dimen100=#1
		\edef\@p@srwidth{\number\dimen100}
}
\def\@p@@ssilent#1{ 
		\@verbosefalse
}
\def\@p@@sprolog#1{\@prologfiletrue\def\@prologfileval{#1}}
\def\@p@@spostlog#1{\@postlogfiletrue\def\@postlogfileval{#1}}
\def\@cs@name#1{\csname #1\endcsname}
\def\@setparms#1=#2,{\@cs@name{@p@@s#1}{#2}}
%
%
\def\ps@init@parms{
		\@bbllxfalse \@bbllyfalse
		\@bburxfalse \@bburyfalse
		\@heightfalse \@widthfalse
		\@scalefalse
		\@rheightfalse \@rwidthfalse
		\def\@p@sbbllx{}\def\@p@sbblly{}
		\def\@p@sbburx{}\def\@p@sbbury{}
		\def\@p@sheight{}\def\@p@swidth{}
		\def\@p@sscale{}
		\def\@p@srheight{}\def\@p@srwidth{}
		\def\@p@sfile{}
		\def\@p@scost{10}
		\def\@sc{}
		\@prologfilefalse
		\@postlogfilefalse
		\@clipfalse
		\if@noisy
			\@verbosetrue
		\else
			\@verbosefalse
		\fi
}
%
%
\def\parse@ps@parms#1{
	 	\@psdo\@psfiga:=#1\do
		   {\expandafter\@setparms\@psfiga,}}
%
%
\newif\ifno@bb
\newif\ifnot@eof
\newread\ps@stream
\def\bb@missing{
	\if@verbose{
		\typeout{psfig: searching \@p@sfile \space  for bounding box}
	}\fi
	\openin\ps@stream=\@p@sfile
	\no@bbtrue
	\not@eoftrue
	\catcode`\%=12
	\loop
		\read\ps@stream to \line@in
		\global\toks200=\expandafter{\line@in}
		\ifeof\ps@stream \not@eoffalse \fi
		\@bbtest{\toks200}
		\if@bbmatch\not@eoffalse\expandafter\bb@cull\the\toks200\fi
	\ifnot@eof \repeat
	\catcode`\%=14
}	
\catcode`\%=12
\newif\if@bbmatch
\def\@bbtest#1{\expandafter\@a@\the#1
\long\def\@a@#1
\long\def\bb@cull#1 #2 #3 #4 #5 {
	\dimen100=#2 bp\edef\@p@sbbllx{\number\dimen100}
	\dimen100=#3 bp\edef\@p@sbblly{\number\dimen100}
	\dimen100=#4 bp\edef\@p@sbburx{\number\dimen100}
	\dimen100=#5 bp\edef\@p@sbbury{\number\dimen100}
	\no@bbfalse
}
\catcode`\%=14
\def\compute@bb{
		\no@bbfalse
		\if@bbllx \else \no@bbtrue \fi
		\if@bblly \else \no@bbtrue \fi
		\if@bburx \else \no@bbtrue \fi
		\if@bbury \else \no@bbtrue \fi
		\ifno@bb \bb@missing \fi
		\ifno@bb \typeout{FATAL ERROR: no bb supplied or found}
			\no-bb-error
		\fi
		\count203=\@p@sbburx
		\count204=\@p@sbbury
		\advance\count203 by -\@p@sbbllx
		\advance\count204 by -\@p@sbblly
		\edef\@bbw{\number\count203}
		\edef\@bbh{\number\count204}
}
%
%
\def\in@hundreds#1#2#3{\count240=#2 \count241=#3
		     \count100=\count240	
		     \divide\count100 by \count241
		     \count101=\count100
		     \multiply\count101 by \count241
		     \advance\count240 by -\count101
		     \multiply\count240 by 10
		     \count101=\count240	
		     \divide\count101 by \count241
		     \count102=\count101
		     \multiply\count102 by \count241
		     \advance\count240 by -\count102
		     \multiply\count240 by 10
		     \count102=\count240	
		     \divide\count102 by \count241
		     \count200=#1\count205=0
		     \count201=\count200
			\multiply\count201 by \count100
		 	\advance\count205 by \count201
		     \count201=\count200
			\divide\count201 by 10
			\multiply\count201 by \count101
			\advance\count205 by \count201
		     \count201=\count200
			\divide\count201 by 100
			\multiply\count201 by \count102
			\advance\count205 by \count201
		     \edef\@result{\number\count205}
}
\def\compute@wfromh{
		\in@hundreds{\@p@sheight}{\@bbw}{\@bbh}
		\edef\@p@swidth{\@result}
}
\def\compute@hfromw{
		\in@hundreds{\@p@swidth}{\@bbh}{\@bbw}
		\edef\@p@sheight{\@result}
}
\def\compute@wfroms{
		\in@hundreds{\@p@sscale}{\@bbw}{100}
		\edef\@p@swidth{\@result}
}
\def\compute@hfroms{
		\in@hundreds{\@p@sscale}{\@bbh}{100}
		\edef\@p@sheight{\@result}
}
\def\compute@handw{
		\if@scale
			\compute@wfroms
			\compute@hfroms
		\else
			\if@height 
				\if@width
				\else
					\compute@wfromh
				\fi	
			\else 
				\if@width
					\compute@hfromw
				\else
					\edef\@p@sheight{\@bbh}
					\edef\@p@swidth{\@bbw}
				\fi
			\fi
		\fi
}
\def\compute@resv{
		\if@rheight \else \edef\@p@srheight{\@p@sheight} \fi
		\if@rwidth \else \edef\@p@srwidth{\@p@swidth} \fi
}
%
\def\compute@sizes{
	\compute@bb
	\compute@handw
	\compute@resv
}
%
%
\def\psfig#1{\vbox {
	%
	\ps@init@parms
	\parse@ps@parms{#1}
	\compute@sizes
	\ifnum\@p@scost<\@psdraft{
		\if@verbose{
			\typeout{psfig: including \@p@sfile \space }
		}\fi
		\special{ps::[begin] 	\@p@swidth \space \@p@sheight \space
				\@p@sbbllx \space \@p@sbblly \space
				\@p@sbburx \space \@p@sbbury \space
				startTexFig \space }
		\if@clip{
			\if@verbose{
				\typeout{(clip)}
			}\fi
			\special{ps:: doclip \space }
		}\fi
		\if@prologfile
		    \special{ps: plotfile \@prologfileval \space } \fi
		\special{ps: plotfile \@p@sfile \space }
		\if@postlogfile
		    \special{ps: plotfile \@postlogfileval \space } \fi
		\special{ps::[end] endTexFig \space }
		\vbox to \@p@srheight true sp{
			\hbox to \@p@srwidth true sp{
				\hss
			}
		\vss
		}
	}\else{
		\vbox to \@p@srheight true sp{
		\vss
			\hbox to \@p@srwidth true sp{
				\hss
				\if@verbose{
					\@p@sfile
				}\fi
				\hss
			}
		\vss
		}
	}\fi
}}
\def\psglobal{\typeout{psfig: PSGLOBAL is OBSOLETE; use psprint -m instead}}
\catcode`\@=12\relax

\section{Introduction}

Lexicalized Tree Adjoining Grammars (LTAG) and Combinatory Categorial
Grammar (CCG) \cite{steedman:97} are known to be weakly
equivalent but not strongly equivalent.  Coordination schema have a
natural description in CCG, while these schema have no natural
equivalent in a standard LTAG.

In \cite{Joshi91} it was shown that in principle it is possible to
construct a CCG-like account for coordination in the framework of
LTAGs, but there was no clear notion of what the derivation structure
would look like. In this paper, continuing the work of \cite{Joshi91},
we show that an account for coordination can be constructed using the
derivation structures in an LTAG.

Using the notions given in this paper we also discuss the construction
of practical parser for LTAGs that can handle coordination including
cases of non-constituent coordination. This approach has been
implemented in the XTAG system \cite{xtagrpt95} thus extending it to
handle coordination.  This is the first full implementation of
coordination in the LTAG framework.

\section{LTAG} 

An LTAG is a set of trees ({\em elementary trees}) which have at least
one terminal symbol on its frontier called the {\em anchor}. Each node
in the tree has a unique address obtained by applying a Gorn tree
addressing scheme, shown in the tree {\em $\alpha$(cooked)}
(Fig.~\ref{fig:ltag}). Trees can be rewritten using {\em
  substitution\/} and {\em adjunction}.  A history of these operations
on elementary trees in the form of a {\em derivation tree} can be used
to reconstruct the derivation of a string recognized by a LTAG.  In
Fig.~\ref{fig:ltag}, the tree {\em $\beta$(dried)} adjoins into {\em
  $\alpha$(beans)} and trees {\em $\alpha$(John)} and {\em
  $\alpha$(beans)} substitutes into {\em $\alpha$(cooked)} to give a
derivation tree for {\em John cooked dried beans}. Each node in the
derivation tree is the name of an elementary tree. The labels on the
edges denote the address in the parent node where a substitution or
adjunction has occured.

\begin{figure}[htbp]
  \begin{center}
    \leavevmode
    \psfig{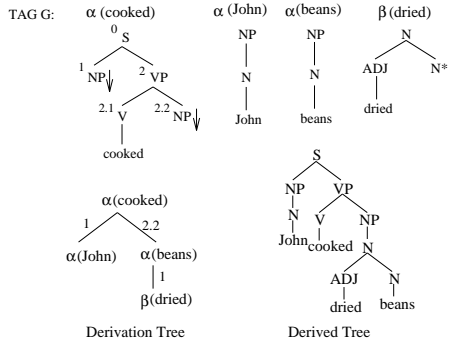}
  \end{center}
  \caption{Example of an LTAG and an LTAG derivation}
  \label{fig:ltag}
\end{figure}
\vspace{-0.12in}
\section{Trees as Structured Categories}

In \cite{Joshi91} elementary trees as well as derived trees in an LTAG
were considered as structured categories defined as a 3-tuple of an
elementary or derived tree, the string it spanned and the functional
type of the tree, e.g $\langle \sigma_1, l_1, \tau_1 \rangle$ in
Fig.~\ref{fig:eats-cookies}. Functional types for trees could be
thought of as defining un-Curried functions corresponding to the
Curried CCG counterpart. A functional type was given to sequences of
lexical items in trees even when they were not contiguous; i.e.
discontinuous constituents were also assigned types.  They were,
however, barred from coordinating.

\begin{figure}[htbp]
  \begin{center}
    \leavevmode
    \psfig{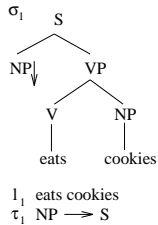}
  \end{center}
  \caption{Structured Category for {\em eats cookies}}
  \label{fig:eats-cookies}
\end{figure}
\vspace{-0.12in}
Coordination of two structured categories $\sigma_1, \sigma_2$
succeeded if the lexical strings of both categories were contiguous,
the functional types were identical, and the least nodes dominating
the strings spanned by the component tree have the same label. For
example, in Fig.~\ref{fig:eats-and-drinks} the tree corresponding to
{\em eats cookies and drinks beer} would be obtained by:
\begin{enumerate}
\item equating the {\em NP} nodes%
  \footnote{ This notion of sharing should not be confused with a
  deletion type analysis of coordination. The scheme presented in
  \cite{Joshi91} as well as the analysis presented in this paper are
  not deletion analyses.}%
\ in $\sigma_1$ and $\sigma_2$, preserving the linear precedence of
the arguments.
\item coordinating the {\em VP} nodes, which are the least nodes
  dominating the two contiguous strings.
\item collapsing the supertrees above the {\em VP} node.
\item selecting the leftmost {\em NP} as the lexical site for the
  argument, since precedence with the verb is maintained by this
  choice.
\end{enumerate}

\begin{figure}[htbp]
  \begin{center}
    \leavevmode
    \psfig{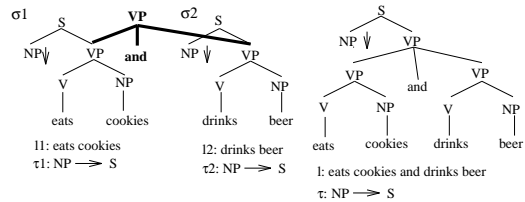}
  \end{center}
  \caption{Coordination of {\em eats cookies and drinks beer}}
  \label{fig:eats-and-drinks}
\end{figure}
\vspace{-0.12in}
The process of coordination built a new derived structure given
previously built pieces of derived structure (or perhaps elementary
structures). There is no clear notion of a derivation structure for
this process.

\section{Coordination in TAG}

An account for coordination in a standard LTAG cannot be given without
introducing a notion of sharing of arguments in the two lexically
anchored trees because of the notion of {\em locality} of arguments in
LTAG. In~\ref{ex:rnr} for instance, the NP {\em the beans} in the
``right node raising'' construction has to be shared by the two
elementary trees (anchored by {\em cooked} and {\em ate}
respectively).

\beginsentences
\sitem (((Harry cooked) and (Mary ate)) the beans)
\label{ex:rnr}
\endsentences

We introduce a notation that will enable us to talk about this more
formally. In Fig.~\ref{fig:ltag} the notation $\downarrow$ denotes
that a node is a non-terminal and hence expects a substitution
operation to occur.  The notation $*$ marks the foot node of an
auxiliary tree.  Making this explicit we can view an elementary tree
as a ordered pair of the tree structure and a ordered
set%
\footnote{ The ordering is given by the fact that the elements of the
  set are Gorn addresses.}%
\ of such nodes from its frontier%
\footnote{ We shall assume there are no adjunction constraints in
  this paper. }%
, e.g. the tree for {\em cooked} will be represented as $\langle
\alpha(cooked), \{ 1, 2.2 \} \rangle$. Note that this representation
is not required by the LTAG formalism. The second projection of this
ordered pair is used here for ease of explication.  Let the second
projection of the pair minus the foot nodes be the {\em substitution
  set}. We will occasionally use the first projection of the
elementary tree to refer to the ordered pair.

{\em Setting up Contractions.} We introduce an operation called {\em
  build-contraction} that takes an elementary tree, places a subset
from its second projection into a {\em contraction set} and assigns
the difference of the set in the second projection of the original
elementary tree and the contraction set to the second projection of
the new elementary tree. The contents of the contraction set of a tree
can be inferred from the contents of the set in the second projection
of the elementary tree. Hence, while we refer to the contraction set
of an elementary tree, it does not have to be stored along with its
representation.

Fig.~\ref{fig:contract-list} gives some examples; each node in the
contraction set is circled in the figure. In the tree $\langle
\alpha(cooked), \{ 1, 2.2 \} \rangle$ application of the operation on
the {\em NP} node at address $2.2$ gives us a tree with the
contraction set $\{ 2.2 \}$. The new tree is denoted by $\langle
\alpha(cooked)_{\{ 2.2 \}}, \{ 1 \} \rangle$, or $\alpha(cooked)_{\{
  2.2 \}}$ for short.  Placing the {\em NP} nodes at addresses $1$ and
$2.2$ of the tree $\alpha(cooked)$ into the contraction set gives us
$\alpha(cooked)_{\{ 1, 2.2 \}}$.

\begin{figure}[htbp]
  \begin{center}
    \leavevmode
    \psfig{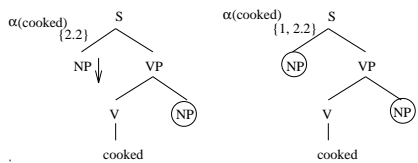}
  \end{center}
  \caption{Building contraction sets}
  \label{fig:contract-list}
\end{figure}

We assume that the anchor cannot be involved in a {\em
  build-contraction}. This assumption needs to be revised when gapping
is considered in this framework~(\S\ref{sec:c-anchor}).

{\em The Coordination Schema.} We use the standard notion of
coordination shown in Fig.~\ref{fig:conj} which maps two constituents
of {\em like type}, but with different interpretations,
into a constituent of the same type%
\footnote{ In this paper, we do not consider coordination of unlike
  categories, e.g. {\em Pat is a Republican and proud of
    it}. \cite{tr:coord} discusses such cases, following \newcite{ja:92}.}%
.

\begin{figure}[htbp]
  \begin{center}
    \leavevmode
    \psfig{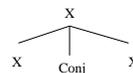}
  \end{center}
  \caption{Coordination schema}
  \label{fig:conj}
\end{figure}
\vspace{-0.12in}
We add a new rewriting operation to the LTAG formalism called {\em conjoin\/}%
\footnote{ Later we will discuss an alternative which replaces this
  operation by the traditional operations of substitution and
  adjunction. }%
. While substitution and adjunction take two trees to give a derived
tree, {\em conjoin\/} takes three trees and composes them to give a
derived tree.  One of the trees is always the tree obtained by
specializing the schema in Fig.~\ref{fig:conj} for a particular
category%
\footnote{ The tree obtained will be a lexicalized tree, with the
  lexical anchor as the conjunction: {\em and}, {\em but}, etc. }%
.

Informally, the conjoin operation works as follows: The two trees
being coordinated are substituted into the conjunction tree. This
notion of substitution differs from the traditional LTAG substitution
operation in the following way: In LTAG substitution, always the root
node of the tree being substituted is identified with the substitution
site. In the conjoin operation however, the node substituting into the
conjunction tree is given by an algorithm, which we shall call {\em
  FindRoot} that takes into account the contraction sets of the two
trees. {\em FindRoot} returns the lowest node that dominates all nodes
in the substitution set of the elementary tree%
\footnote{ This ensures the node picked by {\em FindRoot} always
  dominates a contiguous string in a derivation. This captures the
  string contiguity condition that was used in~\cite{Joshi91}.  A
  coordinated node will never dominate multiple foot nodes.  Such a
  case occurs, e.g., two auxiliary trees with substitution nodes at
  the same tree address are coordinated with only the substitution
  nodes in the contraction set. }%
, e.g. $FindRoot(\alpha(cooked)_{\{ 2.2 \}})$ will return the root
node, i.e.  corresponding to the {\em S conj S} instantiation of the
coordination schema.  $FindRoot(\alpha(cooked)_{\{ 1, 2.2 \}})$ will
return node address $2.1$, corresponding to the {\em V conj V}
instantiation.

The conjoin operation then creates a {\em contraction\/} between nodes
in the contraction sets of the trees being coordinated.  The term {\em
  contraction\/} is taken from the graph-theoretic notion of edge
contraction. In a graph, when an edge joining two vertices is
contracted, the nodes are merged and the new vertex retains edges to
the union of the neighbors of the merged vertices%
\footnote{ Merging in the graph-theoretic definition of contraction
  involves the identification of two previously distinct nodes.  In
  the process of contraction over nodes in elementary trees it is the
  operation on that node (either substitution or adjunction) that is 
  identified. }%
. The conjoin operation supplies a new edge between each corresponding
node in the contraction set and then contracts that edge. As a
constraint on the application of the conjoin operation, the
contraction sets of the two trees must be identical.

Another way of viewing the conjoin operation is as the construction of
an auxiliary structure from an elementary tree. For example, from the
elementary tree $\langle \alpha(drinks), \{ 1, 2.2 \} \rangle$, the
conjoin operation would create the auxiliary structure $\langle
\beta(drinks)_{\{ 1 \}}, \{ 2.2 \} \rangle$ shown in
Fig.~\ref{fig:aux-conj}. The adjunction operation would now be
responsible for creating contractions between nodes in the contraction
sets of the two trees supplied to it. Such an approach is attractive
for two reasons. First, it uses only the traditional operations of
substitution and adjunction. Secondly, it treats {\em conj X} as a
kind of ``modifier'' on the left conjunct {\em X}. We do not choose
between the two representations but continue to view the conjoin
operation as a part of our formalism.

\begin{figure}[htbp]
  \begin{center}
    \leavevmode
    \psfig{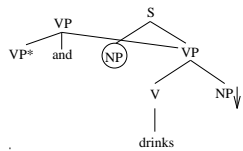}
  \end{center}
  \caption{Coordination as adjunction.}
  \label{fig:aux-conj}
\end{figure}
\vspace{-0.12in}
For example, applying {\em conjoin\/} to the trees {\em Conj(and)},
$\alpha(eats)_{\{1 \}}$ and $\alpha(drinks)_{\{1 \}}$ gives us the
derivation tree and derived structure for the constituent in
\ref{ex:vpc} shown in Fig.~\ref{fig:vpc}.

\beginsentences
\sitem \ldots eats cookies and drinks beer. \label{ex:vpc}
\endsentences

\begin{figure}[htbp]
  \begin{center}
    \leavevmode
    \psfig{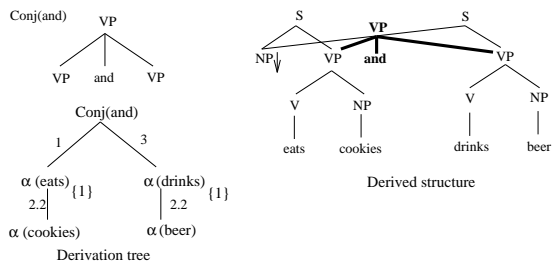}
  \end{center}
  \caption{An example of the {\em conjoin\/} operation.}
  \label{fig:vpc}
\end{figure}

In Fig.~\ref{fig:vpc} the nodes $\alpha(eats)_{\{1 \}}$ and
$\alpha(drinks)_{\{1 \}}$ signify an operation left incomplete at
address $1$.

{\em The Effects of Contraction.} One of the effects of contraction is
that the notion of a derivation tree for the LTAG formalism has to be
extended to an acyclic {\em derivation graph\/}%
\footnote{ We shall use the general notation {\em derivation
    structure} to refer to both derivation trees and
derivation graphs. }%
. Simultaneous substitution or adjunction modifies a derivation tree
into a graph as can be seen in Fig.~\ref{fig:chapman}.

If a contracted node in a tree (after the conjoin operation) is a
substitution node, then the argument is recorded as a substitution
into the two elementary trees as for example in the
sentences~\ref{ex:chapman} and \ref{ex:keats-chapman}.

\beginsentences
\sitem Chapman eats cookies and drinks beer. \label{ex:chapman}
\sitem Keats steals and Chapman eats apples. \label{ex:keats-chapman}
\endsentences

Fig.~\ref{fig:chapman} contains the derivation and derived structures
for \ref{ex:chapman} and Fig.~\ref{fig:keats-chapman} for
\ref{ex:keats-chapman}. In Fig.~\ref{fig:keats-chapman} the derivation
graph for sentence~\ref{ex:keats-chapman} accounts for the
coordinations of the traditional nonconstituent ``Keats steals'' by
carrying out the coordination at the root, i.e. {\em S conj S}. No
constituent corresponding to ``Keats steals'' is created in the
process of coordination.

\begin{figure}[htbp]
  \begin{center}
    \leavevmode
    \psfig{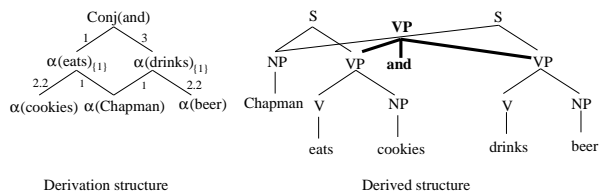}
  \end{center}
  \caption{Derivation for {\em Chapman eats cookies and drinks beer.}}
  \label{fig:chapman}
\end{figure}

\begin{figure}[htbp]
  \begin{center}
    \leavevmode
    \psfig{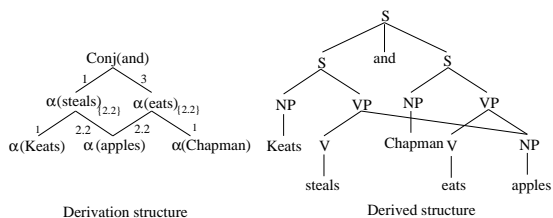}
  \end{center}
  \caption{Derivation for {\em Keats steals and Chapman eats apples.}}
  \label{fig:keats-chapman}
\end{figure}
\vspace{-0.12in}
The derived structures in Figs.~\ref{fig:chapman}
and~\ref{fig:keats-chapman} are difficult to reconcile with traditional
notions of phrase structure%
\footnote{ \newcite{mccawley:82} raised the heterodox view that a
  discontinuous constituent structure should be given for right node
  raising cases, having the same notion of constituency as our
  approach. However, no conditions on the construction of such a
  structure was given. In fact, his mechanism also covered cases of
  parenthetical placement, scrambling, relative
  clause extraposition and heavy NP shift.}%
.  However, the derivation structure gives us all the information
about dependency that we need about the constituents.  The derivation
encodes exactly how particular elementary trees are put together.
Obtaining a tree structure from a derived structure built by the
conjoin operation is discussed in~\cite{tr:coord}.

Considerations of the locality of movement phenomena and its
representation in the LTAG formalism \cite{kj86} can also now explain
constraints on coordinate structure, such as across-the-board
exceptions to the well known coordinate structure constraint, see
Fig.~\ref{fig:atb}. Also in cases of unbounded right node raising such
as {\em Keats likes and Chapman thinks Mary likes beans}, {\em Chapman
  thinks} simply adjoins into the right conjunct of the coordinate
structure%
\footnote{ A comparision of this paper's approach with the
  derivational machinery in CCG and the devices of 3-D coordination is
  done in \cite{tr:coord}.}%
.

\begin{figure}[htbp]
  \begin{center}
    \leavevmode
    \psfig{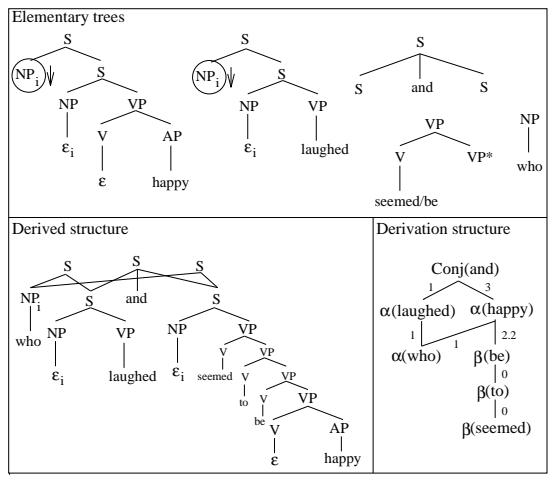}
  \end{center}
  \caption{Derivation for {\em Who laughed and seemed to be happy?}}
  \label{fig:atb}
\end{figure}

\section{Contractions on Anchors}
\label{sec:c-anchor}

An LTAG along with the operations of substitution and adjunction also
has the implicit operation of lexical insertion (represented as the
diamond mark in Fig.~\ref{fig:lexicalization}).  Under this view, the
LTAG trees are taken to be templates.  For example, the tree in
Fig.~\ref{fig:lexicalization} is now represented as $\langle
\alpha(eat), \{ 1, 2.1, 2.2 \} \rangle$.

\begin{figure}[htbp]
  \begin{center}
    \leavevmode
    \psfig{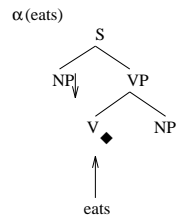}
  \end{center}
  \caption{Lexicalization in a LTAG.}
  \label{fig:lexicalization}
\end{figure}
\vspace{-0.12in}
If we extend the notion of contraction in the conjoin operation
together with the operation of lexical insertion we have the following
observations: The two trees to be used by the conjoin operation are no
longer strictly lexicalized as the label associated with the diamond
mark is a preterminal. Previous uses of conjoin applied to two
distinct trees.  If the lexicalization operation is to apply
simultaneously, the same anchor projects two elementary trees from the
lexicon. The process of contraction ensures that the anchor is placed
into a pair of LTAG tree templates with a single lexical insertion.

{\em Gapping.} Using this extension to {\em conjoin}, we can handle
sentences that have the ``gapping'' construction like
sentence~\ref{ex:gap}.
\beginsentences
\sitem John ate bananas and Bill strawberries. \label{ex:gap}
\endsentences

The conjoin operation applies to copies of the same elementary tree
when the lexical anchor is in the contraction set. For example, let
$\alpha(eats)$ be the tree selected by {\em eats}.  The coordination
of $\alpha(eats)_{\{ 2.1 \}}$ with a copy of itself and the subsequent
derivation tree is depicted in Fig.~\ref{fig:gapping}%
\footnote{ In English, following \newcite{ross:70}, the anchor goes to
  the left conjunct. }%
. 

\begin{figure}[htbp]
  \begin{center}
    \leavevmode
    \psfig{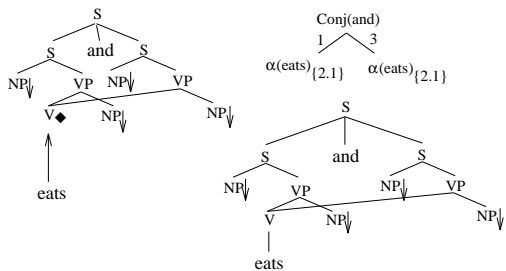}
  \end{center}
  \caption{Handling the gapping construction using contractions.}
  \label{fig:gapping}
\end{figure}

An extension of the approach here will be to permit the conjoin
operation to create contractions on {\em all} the nodes in contraction
sets that it dominates during a derivation, allowing us to recognize
cases of gapping such as: {\em John wants Penn to win and Bill,
  Princeton.} and {\em John wants to try to see Mary and Bill, Susan.}

{\em Coordinating Ditransitive verbs.} In
sentence~\ref{ex:ditrans-conj} if we take the position that the string
{\em Mary a book} is not a constituent (i.e. {\em give} has a
structure as in Fig.~\ref{fig:ditrans}), then we can use the notion
of contraction over the anchor of a tree to derive the sentence
in~\ref{ex:ditrans-conj}. The structure we derive is shown in
Fig.~\ref{fig:ditrans-conj}.

\beginsentences 
\sitem John gave Mary a book and Susan a flower. \label{ex:ditrans-conj}
\endsentences

\begin{figure}[htbp]
  \begin{center}
    \leavevmode
    \psfig{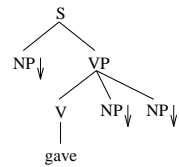}
  \end{center}
  \caption{Tree for a ditransitive verb in LTAG.}
  \label{fig:ditrans}
\end{figure}

\begin{figure}[htbp]
  \begin{center}
    \leavevmode
    \psfig{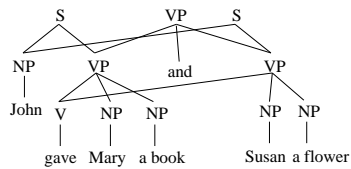}
  \end{center}
  \caption{Derived tree for {\em John gave Mary a book and Susan a flower}.}
  \label{fig:ditrans-conj}
\end{figure}
\vspace{-0.12in}
{\em Interactions.} Permitting contractions on multiple substitution
and adjunction sites along with contractions on the anchor allow the
derivation of {\em sluicing} structures such as~\ref{ex:sluicing}
(where the conjunct {\em Bill too} can be interpreted as {\em [John
  loves] Bill too} or as {\em Bill [loves Mary] too}%
\footnote{ Whether this should be derived syntactically is
  controversial, for example, see~\cite{steedman90}.
 }%
. 

\beginsentences 
\sitem John loves Mary and Bill too. \label{ex:sluicing}
\endsentences

\section{Parsing Issues}

This section discusses parsing issues that arise in the modified TAG
formalism that we have presented. We do not discuss general issues in
parsing TAGs, rather we give the appropriate modifications that are
needed to the existing Earley-type parsing algorithm for TAGs due to
\newcite{Schabes88a}.

The algorithm relies on a tree traversal that scans the input string
from left to right while recognizing the application of the conjoin
operation. The nodes in the elementary trees are visited in a top-down
left to right manner (Fig.~\ref{fig:tree-traversal}). Each dot in
Fig.~\ref{fig:tree-traversal} divides the tree into a left context and
a right context, enabling the algorithm to scan the elementary tree
while trying to recognize possible applications of the conjoin
operation.

\begin{figure}[htbp]
  \begin{center}
    \leavevmode
    \psfig{figure=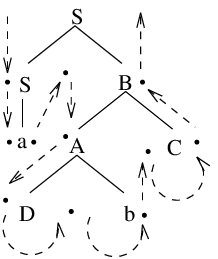,scale=65}
    \caption{Example of a tree traversal}
    \label{fig:tree-traversal}
  \end{center}
\end{figure}
\vspace{-0.12in}
The derived structure corresponding to a coordination is a composite
structure built by applying the conjoin operation to two elementary
trees and an instantiation of the coordination schema. The algorithm
never builds derived structures. It builds the derivation by visiting
the appropriate nodes during its tree traversal in the following order
(see Fig.~\ref{fig:recognize-conj}).

\[ 1\ 2 \cdots 3\ 4 \cdots 5\ 6 \cdots 2'\ 7' \cdots 3'\ 4' \cdots 5'\ 6' \cdots 7\ 8 \]

The algorithm must also compute the correct span of the string for the
nodes that have been identified via a contraction.
Fig.~\ref{fig:recognize-conj} gives the possible scenarios for the
position of nodes that have been linked by a contraction.  When foot
nodes undergo contraction, the algorithm has to ensure that both the
foot nodes share the subtree pushed under them, e.g. $9 \cdots 10$ and
$9' \cdots 10'$ in Fig.~\ref{fig:recognize-conj}(a).  Similarly, when
substitution nodes undergo contraction, the algorithm has to ensure
that the tree recognized due by predicting a substitution is shared by
the nodes, e.g. $11 \cdots 12$ and $11' \cdots 12'$ in
Figs.~\ref{fig:recognize-conj}(b) and~\ref{fig:recognize-conj}(c). The
traversals $9 \cdots 10$ should span the same length of the input as
$9' \cdots 10'$, similarly for $11 \cdots 12$ and $11' \cdots 12'$.
Various positions for such traversals is shown in
Fig.~\ref{fig:recognize-conj}. A derivation is valid if the input
string is accepted and each node in a contraction spans a valid
substring in the input. The complete and formal description of the
parsing algorithm is given in~\cite{tr:coord}.

\begin{figure}[htbp]
  \begin{center}
    \leavevmode \psfig{figure=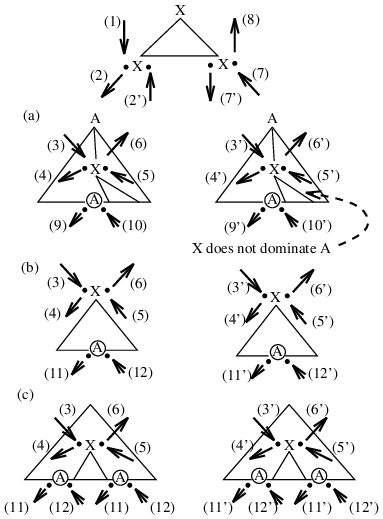,scale=80}
    \caption{Moving the dot while recognizing a conjoin operation}
    \label{fig:recognize-conj}
  \end{center}
\end{figure}

\section{Conclusion}

We have shown that an account for coordination can be given in a LTAG
while maintaining the notion of a derivation structure which is
central to the LTAG approach. We showed that fixed constituency can be
maintained at the level of the elementary tree while accounting for
cases of non-constituent coordination and gapping. We discussed the
construction of a practical parser for LTAG that can handle these
cases of coordination.

\end{document}